# An Evolving Cascade Neural Network Technique for Cleaning Sleep Electroencephalograms

VITALY SCHETININ

*Computer Science Department, University of Exeter, Exeter, EX4 4QF, UK*

V.Schetinin@ex.ac.uk

**Abstract.** Evolving Cascade Neural Networks (ECNNs) and a new training algorithm capable of selecting informative features are described. The ECNN initially learns with one input node and then evolves by adding new inputs as well as new hidden neurons. The resultant ECNN has a near minimal number of hidden neurons and inputs. The algorithm is successfully used for training ECNN to recognise artefacts in sleep electroencephalograms (EEGs) which were visually labelled by EEG-viewers. In our experiments, the ECNN outperforms the standard neural-network as well as evolutionary techniques.

***Key words:*** *neural network, cascade architecture, evolving, feature selection, electroencephalogram*

## 1. Introduction

To build feed-forward neural networks, a cascade-correlation learning algorithm [1, 2] has been suggested which generates hidden neurons as they are needed. Several authors have explored and applied cascade neural networks to real-world problems [3 - 5]. A cascade network differs from fully connected feed-forward neural networks (FNNs) exploiting a fixed architecture – in contrast to the FNNs, cascade networks start learning with only one neuron, and during learning the algorithm automatically adds and trains new neurons creating a multi-layer structure. The number of hidden neurons, that is the complexity of the network, increases step-by-step while the training error decreases. As a result, the training algorithm grows the neural network of a near optimal complexity which can generalise well.

In [4], Phatak and Koren have modified the algorithm to generate networks with restricted fan-in and a small number of hidden layers (or depth) by controlling the connectivity. Their results reveal that there is a trade-off between connectivity and depth, number of independent parameters, learning time, *etc.* When the number of inputs is small relative to the size of training set, a higher connectivity usually leads to faster learning and fewer independent parameters, but it also results in unbounded fan-in and depth.

In practice, training algorithms can over-fit cascade networks because of noise in training data. Noise affects the features, assumed to be presenting the training data, and makes some of them irrelevant to the classification problem. To overcome the over-fitting problem, data pre-processing techniques have been developed aimed to select the informative features [5, 6]. However, the results of these feature selection algorithms depend on some special conditions, for example, on the order in which the features are processed [7].

To prevent the cascade neural networks from over-fitting, Tetko *et al.* [8] have suggested a method based on a combination of two algorithms, early stopping and ensemble averaging. They have shown that their method improves the predictive ability of the cascade networks.

The pruning methods described in [9] have been developed for networks trained by the cascade-correlation learning algorithm. These methods were used to estimate the importance of the input variables characterising the quantitative relationships. The cascade-correlation networks were compared with the neural networks of a fixed structure in terms of the performance. The use of the selected input variables has improved the predictive accuracy of the cascade neural networks.

Another approach to learning neural network from data, known as Group Method of Data Handling (GMDH), has been suggested by Ivakhnenko [10]. Using an evolutionary principle, the GMDH is effectively used for training the neural networks with growing architecture [11 - 13]. The GMDH algorithms are capable of selecting the relevant features during learning the neural networks. Within the GMDH approach, the complexity of the neural networks grows until a predefined criterion is met. As a result, the GMDH-type neural networks can achieve a near optimal complexity.

In this paper we mainly focus on selecting the most informative features while the cascade networks learn from data. We believe that this is the most effective way to prevent the networks from over-fitting. Indeed, when the cascade network starts to learn with a small number of inputs and neurons, the new inputs and neurons added the network can improve its performance. As such networks evolve during learning, we name them Evolving Cascade Neural Networks (ECNNs).

For fitting the neuron weights to the training data, we used a modified projection method described in [14]. This method allows the neuron weights to be effectively evaluated in the presence of noise whose structure and parameters are unknown.

In this paper we applied the ECNN technique to automatically recognise artefacts in clinical electroencephalograms (EEG) recorded from newborns during sleep hours. These EEG data are characterises by features calculated in the spectral domain. However, some of these features are irrelevant or redundant which makes the recognition problem difficult [15-18].

Analysing sleep EEGs, Roberts and Tarassenko [19] have quantitatively investigated the number of different human sleep states using the high order Kalman filter coefficients averaged over a one-second window. These coefficients were then clustered with a Kohonen self-organizing network. Three types of transition trajectories were found corresponding to states of wakefulness, dreaming sleep, and deep sleep. Following this idea, Schlögl *et. al.* [20], have used inverse filtering to identify artefacts scored by 9 types in sleep EEGs. Kalman filtering was used to estimate adaptive autoregressive parameters. They have concluded that the variance of the prediction error can be used as an indicator for muscle and movement artefacts.

In [21], Roberts *et al.* suggested the technique of detecting outlying patterns existing in real data which may deteriorate the analysis of sleep states. They suggested using an artificial class located outside class boundaries to which a pattern is assigned if the probability of a novelty test is largest.

In our experiments with artefact recognition, first we use a few EEG records whose segments were visually labelled by one EEG-expert. Many more EEG records were used in our second experiments which were labelled by several experts.

Section 2 describes the idea behind the cascade-correlation architecture, and Section 3 describes the ECNN training algorithm we developed. Then Section 4 describes the application of the ECNN algorithm to recognise the artefact segments in the clinical EEGs. Sections 5 and 6 describe the comparison of the ECNN with standard feed-forward neural-network and evolutionary techniques on the EEGs, and finally Section 7 concludes the paper.

## 2. Training Cascade Neural Networks

In this section, we first discuss the cascade neural networks and highlight their main advantages. Second, we describe the algorithm we developed to train ECNNs.

## 2.1. Cascade Neural Networks

The ideas behind the cascade-correlation architecture are as follows. The first is to build up the cascade architecture by adding new neurons together with their connections to all the inputs as well as to the previous hidden neurons. This configuration is not changed at the following layers. The second idea is to learn only the newly created neuron by fitting its weights so that to minimise the residual error of the network. The new neurons are added to the network while its performance increases. So, the common cascade-correlation technique assumes that all $m$ variables $x_1, \ldots, x_m$ characterising the training data are relevant to the classification problem.

At the beginning, a cascade network with $m$ inputs and one output neuron starts to learn without hidden neurons. The output neuron is connected to every input by weights $w_1, \ldots, w_m$ adjustable during learning.

The output $y$ of neurons in the network is given by the standard sigmoid function $f$ as follows

$$y = f(\mathbf{x}; \mathbf{w}) = 1/(1 + \exp(-w_0 - \Sigma_i^m w_i x_i)), \qquad (1)$$

where $\mathbf{x} = (x_1, \ldots, x_m)$ is a $m{\times}1$ input vector, $\mathbf{w} = (w_1, \ldots, w_m)$ is a $m{\times}1$ weight vector and $w_0$ is the bias term which is hereinafter omitted.

Then the new neurons are added to the network one-by-one. Each new neuron is connected to all $m$ inputs as well as to all the previous hidden neurons. Each time only the output neuron is trained. For training, any of algorithms suitable for learning a single-neuron can be used.

Training a new neuron, the algorithm adjusts its weights so that to reduce the residual error of the network. The algorithm adds and then trains the new neurons while the residual error decreases.

The advantages of the cascade neural networks are well known. First, no structure of the networks is predefined, that is, the network is automatically built up from the training data. Second, the cascade network learns fast because each of its neurons is trained independently to each other.

However, a disadvantage is that the cascade networks can be over-fitted in the presence of noisy features. To overcome this problem, we developed a new algorithm for training the ECNN described next.

## 2.2. Evolving Cascade Neural Networks

Let us define a cascade network architecture consisting of neurons whose number of inputs, $p$, is increased from one layer to the next. At the first layer, the neuron is connected to two inputs $x_{i1}, \ldots, x_{i2}, i_1 \neq i_2 \in (1, m)$. Let the input $x_{i1}$ be an input for which a single-input neuron provides a minimal error.

At the second layer, the new neuron is connected with the input $x_{i1}$ as well as with the output of the previous neuron. The third input of this neuron can be connected with that input which provides a maximal decrease in the output error. Each neuron at the new layer can be connected in the same manner.

More formally, the neuron at the $r$th layer has $p = r + 1$ inputs. For a logistic activation function, the output $z_r$ of this neuron can be written as follows

$$z_r = f(\mathbf{u}; \mathbf{w}) = 1/(1 + \exp(-\Sigma_l^p u_l w_l)), \qquad (2)$$

where $r$ is the number of layer, and $\mathbf{u} = (u_1, \ldots, u_p)$ is a $p{\times}1$ input vector of the $r$th neuron.

As an example, Fig. 1 depicts a cascade network for $m = 4$ inputs and $r = 3$ layers. The squares depict the synaptic connections between the output neuron, two hidden neurons with outputs $z_1$ and $z_2$ and the inputs $x_1$, ..., $x_4$.

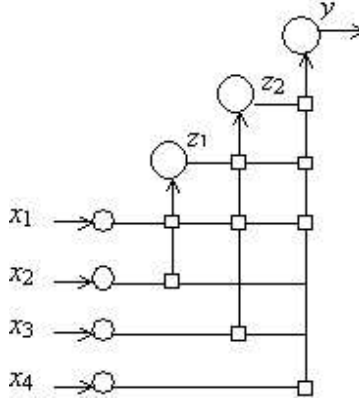

**Fig. 1:** An examples of a trained cascade network with $r = 3$ layers.

So, using the above algorithm, we can estimate in an *ad hoc* manner the decrease in the output error for each feature involved in the combination with the previous features. It is natural to assume that if the output error is evaluated on a validation dataset, the irrelevant as well as redundant features are unlikely to become involved in the resultant network. From this point of view, the selection criterion operates as a regularity criterion in GMDH mentioned above.

The regularity criterion $C_r$ is calculated for the $r$th neuron on the unseen examples, which were not used for fitting the synaptic weights of the neuron. In this case the values of $C_r$ are dependent on the generalisation ability of the neuron with the given connections – the value of $C_r$ increases proportionally to the number of the misclassified validation examples. In other words, the $r$th neuron with irrelevant connections cannot classify all the unseen examples correctly and for this reason the value of $C_r$ is expected to be high.

The idea behind our algorithm is to use the above criterion to select neurons with relevant connections. This criterion says that if the value of $C_r$ calculated for the $r$th neuron is less than the value of $C_{r-1}$ calculated for the previous neuron, then the connection of the $r$th neuron are more relevant than that for the previous layer, else they are less relevant. Formally, this criterion can be used to define the following acceptance rule

$$if\ C_r < C_{r-1},\ then\ \text{accept the } r\text{th neuron, } else \text{ reject it.} \qquad (3)$$

If rule (3) is met, then the connections and the weights of the $r$th neuron are added to the network. In the case, when no neuron is accepted by this rule after the given number of failed attempts, the algorithm stops, and the $r$th neuron with minimal value of $C_r$ is assigned to be the output neuron.

Next, we describe the algorithm for training ECNN in more detail.

## 3. Training of Evolving Cascade Neural Networks

In this section first we describe the method we developed for fitting the weights of neurons. Then, we describe the ECNN training algorithm in detail.

### 3.1. Fitting of Weights

For real-world problems, the structure and parameters of noise affecting data can be unknown. Without this information, the standard evaluation methods, e.g. assuming a Gaussian noise, can yield biased estimates of the weights of neurons. However, a projection method described in [14] can yield unbiased estimates in the presence of noise of the unknown structure. Based on this method, we developed our method of fitting the weights described below.

As mentioned in the previous section, the fitness of the weights of neurons with respect to the training data can be evaluated by the regularity criterion. We can implement this criterion by dividing the whole training dataset $\mathbf{D}$ into two subsets, say, $\mathbf{D}_A$ and $\mathbf{D}_B$, the first for the fitting and the second for validation of the weights. Clearly, in this case the outputs of the neurons with relevant and irrelevant connections calculated on the dataset $\mathbf{D}_B$ have to be significantly different.

Let $\mathbf{D} = (\mathbf{X}, \mathbf{Y}^o)$ be a dataset consisting of $n$ examples, where $\mathbf{X}$ is a $n \times m$-matrix of data, and $\mathbf{Y}^o = (y_1{}^o, \ldots, y_n{}^o)$ is a $n \times 1$ target vector. Correspondingly, we can define $\mathbf{D}_A = (\mathbf{X}_A, \mathbf{Y}_A{}^o)$ and $\mathbf{D}_B = (\mathbf{X}_B, \mathbf{Y}_B{}^o)$ consisting of $n_A$ and $n_B$ examples, respectively, where $n = n_A + n_B$. The user can set the proportion of these subsets, e.g., $n_A = n_B$.

For the ECNN structure defined in Section 2.2, the input $\mathbf{u}_1$ at the first layer is given by two features

$$\mathbf{u}_1 = (x_i, x_{j1}), i \neq j_1 = 1, \ldots, m.$$

At the second layer, the neuron is connected with the firts neuron, the input $x_i$ as well as with the new input $x_{j2}$

$$\mathbf{u}_2 = (z_1, x_i, x_{j2}), j_2 = 1, \ldots, m,$$

where $z_1$ is the output of the first neuron.

Then, at the $r$th layer, the neuron is connected with all the previous neurons and the inputs $x_i$ and $x_{jr}$:

$$\mathbf{u}_r = (z_1, \ldots, z_{r-1}, x_i, x_{jr}), j_r = 1, \ldots, m.$$

where $z_i$ are the outputs of the hidden neurons.

As training and validation of the neuron are realised on different subsets $\mathbf{D}_A$ and $\mathbf{D}_B$, let us denote the neuron inputs as $\mathbf{u}_A$ and $\mathbf{u}_B$, respectively. Then, the $i$th example taken from data $\mathbf{D}$ is $\mathbf{u}^{(i)}$ and $y_i{}^o$.

Firstly, the weight vector $\mathbf{w}^0$ is initialised by random values which can be drawn from a Gaussian distribution $N(0, s_w)$, where $s_w$ is the given variance. Then at the first and further steps $k$, the algorithm calculates the $n_A \times 1$ error vector $\boldsymbol{\eta}_A{}^{(k)}$ on the data $\mathbf{D}_A$ as follows

$$\boldsymbol{\eta}_A{}^{(k)} = f(\mathbf{u}_A, \mathbf{w}^{(k-1)}) - \mathbf{Y}_A{}^o. \tag{4}$$

On the validation data $\mathbf{D}_B$, the vector $\boldsymbol{\eta}_B{}^k$ is calculated as

$$\boldsymbol{\eta}_B{}^{(k)} = f(\mathbf{u}_B, \mathbf{w}^{(k-1)}) - \mathbf{Y}_B{}^o.$$

The residual square error (RSE) $e_B$ of the neuron on the validating dataset is

$$e_B(k) = (\boldsymbol{\eta}_B{}^{(k)} \boldsymbol{\eta}_B{}^{(k)T})^{1/2}.$$

The goal of the fitting algorithm is to adjust the weights $\mathbf{w}$ so that to minimise the value of $e_B$ for a finite number $k^*$ of steps. Clearly, the minimum of $e_B$ is dependent on the level $\varepsilon$ of noise in the data. So, if the noise level is known, we can stop the algorithm when the following condition is met

$$e_B(k^*) \leq \varepsilon \tag{5}$$

For steps $k < k^*$, when the above condition is not met, the current weights are updated by the following learning rule

$$\mathbf{w}^{(k)} = \mathbf{w}^{(k-1)} - \chi \| \mathbf{U}_A \|^{-2} \mathbf{U}_A \mathbf{\eta}_A^{(k-1)}, \tag{6}$$

where $\chi$ is the learning rate, $\mathbf{U}_A$ is the $p \times n_A$ matrix of the input data, $\| \cdot \|$ is a Euclidian norm.

In [14] shown that, when $\chi$ is given between 1 and 2, the procedure (6) always yields the desirable estimates of weights within a given accuracy $\varepsilon$ for a finite number of steps $k^*$.

A simple explanation of the above learning rule can be given in a space of two weight components $w_1$ and $w_2$. Let us assume that in this space there is a desirable region $\mathbf{w}^*$ for which condition (5) is met for any vector $\mathbf{w} \in \mathbf{w}^*$. Assume that at step $k$ a vector $\mathbf{w}^{(k)} \notin \mathbf{w}^*$ as depicted in Fig. 2.

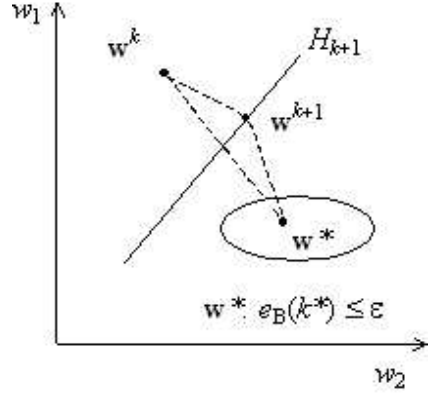

**Fig. 2:** Projection method in a space of two weight components $w_1$ and $w_1$.

Obviously, value $e_B$ is proportional to the distance between the current vector $\mathbf{w}^k$ and the region $\mathbf{w}^*$. Accordingly to rule (6), the new vector $\mathbf{w}^{(k+1)}$ is an orthogonal projection of vector $\mathbf{w}_k$ on hyperplane $H_{k+1}$ located between $\mathbf{w}^k$ and region $\mathbf{w}^*$. We can see that the new vector $\mathbf{w}^{(k+1)}$ is closer to the desirable region $\mathbf{w}^*$ than the previous vector $\mathbf{w}^{(k)}$, and therefore $e_B(k+1) < e_B(k)$. By induction, we can write that for any $k \leq k^*$, $e_B(k) < e_B(k-1) < \ldots < e_B(0)$ is a monotonically decreasing series.

In our experiments we varied $\chi$ from 1.25 to 2.0 and obtained different learning curves shown in Fig. 3. As we can see, the value of RSE is decreased with maximal speed for learning rate $\chi = 2.0$.

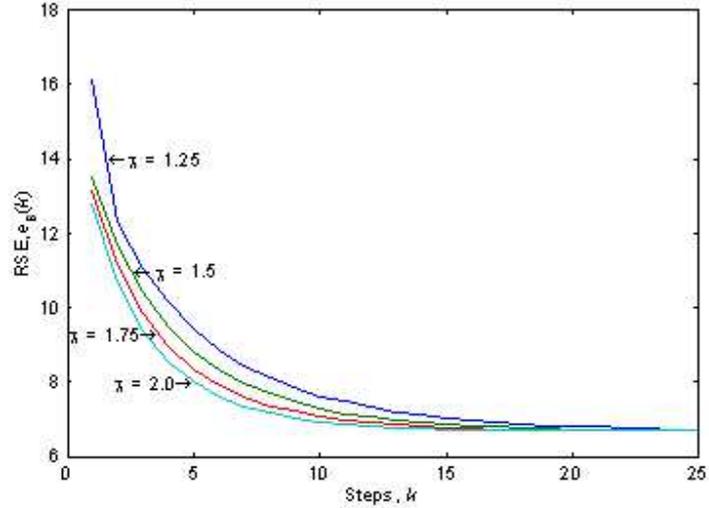

**Fig. 3:** The learning curves were calculated for χ = 1.25, 1.5, 1.75 and 2.0 on the EEG data.

In practice the level of noise in data can be unknown. In this case, instead of rule (5), we can give a constant Δ > 0 defining a minimal increase in the RSE between steps $k - 1$ and $k$. Then we can stop the training algorithm stops if the following rule is met

$$e_B(k*{-}1) \; - \; e_B(k*) \; < \; \Delta. \tag{7}$$

Thus, after $k*$ steps, the algorithm provides a desired weight vector **w**\* for a given constant Δ. Correspondingly, $C_r = e_B(k*)$ is finally used in the acceptance rule (3) introduced to distinguish between the relevant and irrelevant connections.

In our experiments the best performance of ECNN was obtained with χ = 1.9 and Δ = 0.0015. In this case the number $k*$ usually did not exceed 30 steps as depicted in Fig. 3.

### 3.2. The ECNN Training Algorithm

For training ECNN we used the following heuristics. The first is to exploit the best feature which provides a minimal value of $C_r$ calculated for single-input neurons. This feature is connected with each neuron. The second heuristic is to incrementally involve the new features in the cascade network. The third heuristic is to use rule (3) to accept the new neuron if its connections are more relevant than that for the previous neuron.

In order to realise the above heuristics, the ECCN training algorithm includes the following steps.

1.  Initialise the layer $r = 0$ and a set $X := (x_1, \dots, x_m)$. Calculate the values of $S_i = CR_i$ for single-input neurons with one inputs $x_i$, $i = 1, \dots, m$.
2.  Arrange the calculated values $S_i$ in ascending order and put them in a list $S$: $S := \{S_{i1} \leq S_{i2} \leq \dots \leq S_{im}\}$. Set a value $C_0 = S_{i1}$.
3.  Set a position $h = 2$ for the next feature in a set $X$ and a list $S$.
4.  Set $r := r + 1$ and $p = r + 1$. Create the new candidate-neuron with $p$ inputs.

5. If $r > 1$, then connect the first $r$ inputs of this neuron to all previous neurons and to an input $x_{i1}$, respectively. Otherwise, connect this neuron to an input $x_{i1}$.
6. Connect a $p$-th input of a candidate-neuron to an input node being in a position $h$ of a set $X$.
7. Train the candidate-neuron and then calculate its value $C_r$.
8. If $C_r \geq C_{r-1}$, then go to step 10.
9. Put the candidate-neuron to the network as the $r$-th neuron. Go to step 4.
10. If $h < m$, then $h := h + 1$ and go to step 6, else stop.

The weights of candidate-neurons are updated by rule (6) until the condition (7) is met. At step $k = 1$, the neurons start to learn with one input. At the following steps, the cascade network involves new features as well as new neurons while the value $C_r$ decreases.

Finally, the resultant cascade network consists of a near minimal number of connections and neurons. Such networks as we know are able to generalise well.

Next, we describe the application of the ECNN to artefact recognition in the sleep EEGs. These EEG data are characterised by many irrelevant features.

## 4. Cleaning the Sleep Electroencephalograms

The recognition of artefacts in sleep EEGs recorded in clinical conditions is still a difficult problem described by several researchers [15 -18]. For recognising artefacts in EEGs recorded from newborns during sleep hours vie the standard electrodes C3 and C4, Breidbach *et al.* [17] have developed a neural network technique taking in account 72 spectral and statistical features calculated per 10s segment into 6 frequency bands: sub-delta (0-1.5 Hz), delta (1.5-3.5 Hz), theta (3.5-7.5 Hz), alpha (7.5-13.5 Hz), beta 1 (13.5-19.5 Hz), and beta 2 (19.5-25 Hz).

Following [17], in our experiments we used the same structure of EEG data recorded from newborns during sleep hours. The spectral powers of the 6 frequency bands were calculated for channels C3, C4 as well as their sum C3+C4. These features were extended with the statistical features presenting the relative and absolute powers as well as their variances. Finally, the EEG data were normalised to be with zero mean and unit variance.

In our first experiment, we used 2 EEG records in which artefacts were manually labelled by an EEG-viewer. The EEG segments were merged in one dataset and then divided into the training and testing subsets containing 2244 and 1210 randomly selected segments. The rates of EEG artefacts in these datasets were 9.3% and 8.2%, respectively.

Since the initial weights of neurons are randomly assigned, we trained ECNN 100 times. The best on the validation data ECNN misclassified 3.92% of the training and 3.31% of the testing examples.

Fig. 4 depicts a structure of the best ECNN which includes four inputs, three hidden neurons and one output neuron. From the original 72 features, the training algorithm has selected four features $x_{36}$, $x_{23}$, $x_{10}$, and $x_{60}$ which make the most important contribution to the classification outcome. The first hidden neuron is connected to inputs $x_{36}$ and $x_{23}$, and the output neuron is connected with the outputs $z_1$, $z_2$ and $z_3$ of the hidden neurons as well as with inputs $x_{36}$ and $x_{60}$.

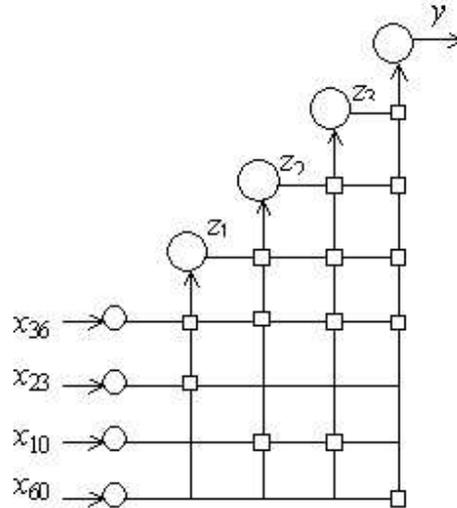

**Fig. 4:** The structure of an ECNN trained to recognise the EEG artefacts.

For the above network, Fig. 5 depicts the correlation coefficients $C_{36}$, $C_{23}$, $C_{10}$ and $C_{60}$ calculated between variables $x_{36}$, $x_{23}$, $x_{10}$ and $x_{60}$ and the remaining variables. We can see that these variables are strongly correlated with some others for which the correlation coefficients are close to 1. This fact explains the variety of structures of ECNNs capable of recognising EEG artefacts with the same effectiveness.

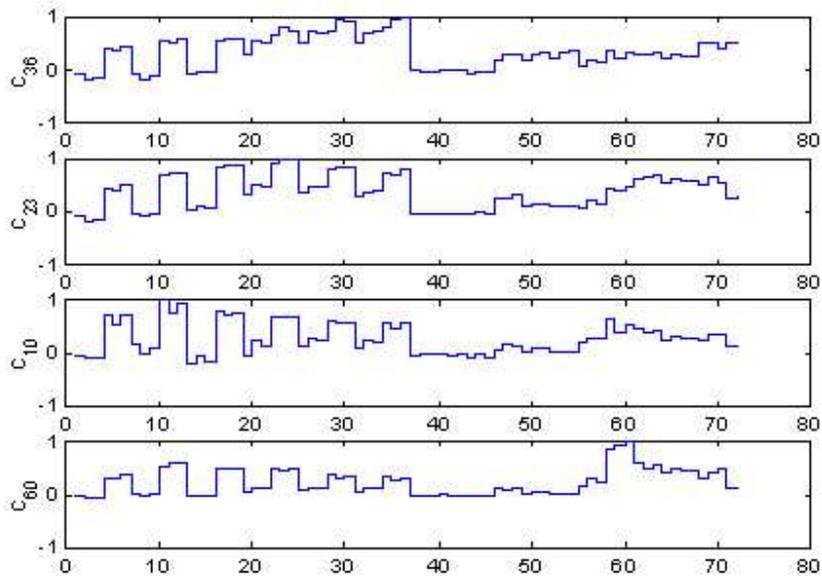

**Fig. 5:** Correlation coefficients $C_{36}$, $C_{23}$, $C_{10}$ and $C_{60}$ calculated between the features $x_{36}$, $x_{23}$, $x_{10}$ and $x_{60}$ and the remaining features.

As the neuron weights calculated by the learning rule (6) are randomly initialised, the resultant weights can slightly vary over the 100 runs. These variations cause the variations in the ECNN structures as well.

Fig. 6 depicts the frequencies of involving the input variables in the ECNNs over the 100 runs. We can see that variable $x_{36}$ is used more frequently than the other input variables. However, rigorously analysing these frequencies, we cannot conclude that they reflect the contribution of the features to the classification outcome. Indeed, the most of the ECNN became stuck in the local maxima of the performance because they missed the best combinations and sequences of the features. Only a few ECNNs have reached a deepest maximum of the performance. One of such ECNNs is depicted in Fig. 4. So, the frequencies of using the variables rather reflect the biased contribution of them.

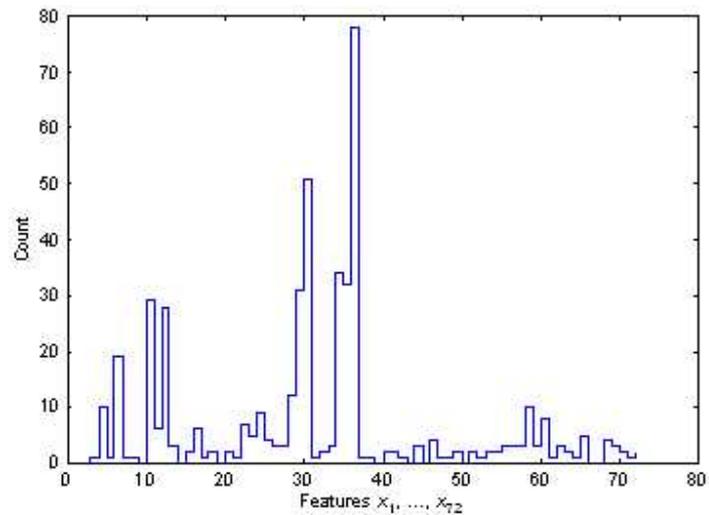

**Fig. 6:** A histogram of the features involved in the ECNNs over 100 runs.

The sizes of the ECNNs over 100 runs vary between 1 and 11 neurons, and the ECNN consisting of four neurons appears with the most frequency. The histogram of the ECNN sizes over 100 runs is depicted in Fig. 7.

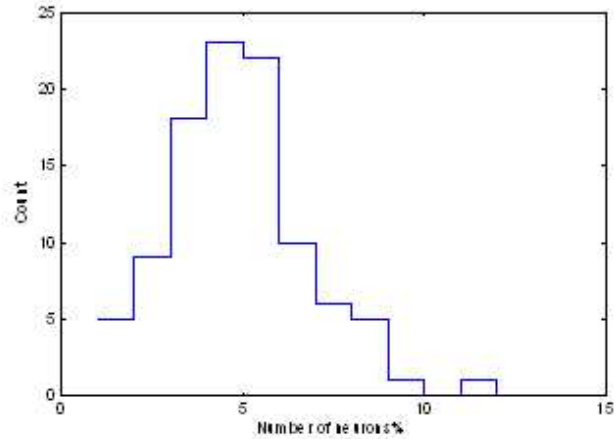

**Fig. 7:** A histogram of the ECNN sizes over 100 runs.

At the same time, the training and testing errors also vary over runs. The histograms of these errors are depicted in Fig. 8 and Fig. 9. We can see that the minimal training and testing error rates are 3.92% and 2.73%, respectively.

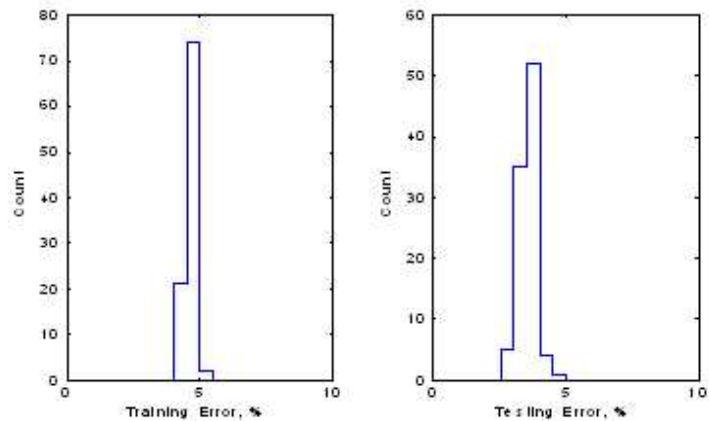

**Fig. 8 and 9:** A histogram of the training (left side) and testing (right side) errors over 100 runs of the ECNN.

## 5. The ECNN versus the Standard Neural Network Technique

For comparison we used the standard neural-network technique to train FNNs with one hidden layer and one output neuron. The number of hidden neurons varied from 2 to 8 neurons. All neurons implemented a standard sigmoid transfer function (1).

In order to remove the contribution of the correlated inputs and improve the performance of the FNNs, we applied the standard technique of Principal Component Analysis (PCA) with different fractions $fr$ of the total variation.

For training the FNNs, we used a fast Levenberg-Marquardt (LM) algorithm provided by MATLAB. The learning parameters of the LM algorithm are listed in Table 1.

**Table 1:** The parameters of the back-propagation learning algorithm

| Net Parameters | Value | Comments |
|---|---|---|
| net.layers | initnw | The hidden and output layers are randomly initialised. Active regions of the layer neurons are distributed roughly evenly over the input space |
| net.trainParam.epochs | 100 | Maximum number of epochs |
| net.trainParam.goal | 0 | Performance goal |
| net.trainParam.lr | 0.01 | Learning rate |
| net.trainParam.max_fail | 5 | Maximum validation failures |
| net.trainParam.min_grad | e-10 | Minimum gradient |
| net.trainParam.mu | 0.001 | Parameter MU of ML algorithm |
| net.trainParam.mu_dec | 0.1 | MU multiplier |
| net.trainParam.mu_inc | 10 | MU multiplier |
| net.trainParam.mu_max | e+10 | Maximum MU |
| net.trainParam.time | Inf | Training time |

To prevent the FNNs from over-fitting, we used the standard early stopping rule, for which the training data have been divided into the training and validation data subsets. The proportions of the data points for these sets were 2/3 and 1/3, respectively.

For each given variant of the PCA, we trained the FNNs with different number of hidden neurons 100 times randomly initialising their weights. The performance of the FNNs was evaluated on the testing data.

As a result, we found the best FNN with 4 hidden neurons and 11 principal components which have been found for $fr = 0.02$. The training and testing error rates were 2.97% and 5.54%, respectively.

In our experiments, the error rates over 100 runs of the FNNs varied between 2.77% and 8.71%. The histograms of the training and testing error rates of the FNN over 100 runs are depicted in Fig. 10 and Fig. 11.

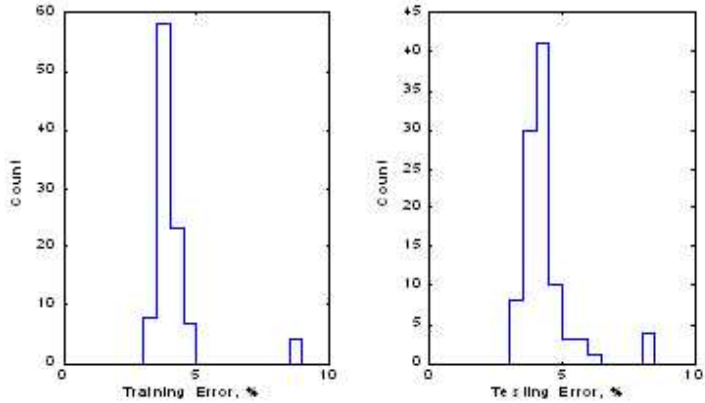

**Fig. 10 and 11:** A histogram of the training (left side) and testing (right side) errors over 100 runs of the FNN.

Comparing the performances of the best ECNN versus the best FNN on the same EEG data, we see that these networks made 3.31% and 5.45%, respectively. Hence, we can conclude that the ECNN slightly out-performs the standard FNN technique.

On the other hand, we found that the standard network using a structure discovered by the ECNN, depicted in Fig. 4, performs even better. In our experiment, the FNN*, which includes three hidden neurons and four inputs $x_{10}$, $x_{23}$, $x_{36}$ and $x_{60}$, misclassified 3.30% on the training and 2.98% on the testing datasets.

Finally, Table 2 presents the training and testing error rates for the best ECNN, FNN, and FNN*. The column Test* provides the minimal testing error rates for these networks over 100 runs.

**Table 2:** The error rates of the ECNN, FNN, and FNN*.

| # | Method | Error rates, % | | |
|---|--------|-------|------|-------|
| | | Train | Test | Test* |
| 1 | *ECNN* | 3.92 | 3.31 | 2.73 |
| 2 | *FNN* | 2.97 | 5.54 | 3.06 |
| 3 | *FNN*\* | 3.30 | **2.98** | 2.73 |

All comparative experiments described above have been carried out with MATLAB 5. The average learning time required for the ECNN did not exceed that for the fast LM algorithm. For the ECNN this time is dependent on the learning parameters χ and Δ specified in (6) and (7).

Thus, we can conclude first that the ECNN slightly outperforms the standard FNN on the same testing data. Second, the best ECNN involves the 4 most relevant features selected from the original 72 features and consists of 4 neurons. That is, the ECNN training algorithm is able to automatically discover the neural network structure appropriate to the training data. Third, the use of the discovered structure for the standard FNN technique allows achieving an increase in the performance as shown for the FNN*. That is, the ECNN algorithm can be used as a preprocessing technique which is more effective than the standard PCA.

## 6. The ECNN versus the Evolutionary and Decision Tree Techniques

In our second comparative experiments, we used EEGs recorded from 30 newborns as described for the first experiments. 20 EEGs containing 21624 segments form the training and validation data sets, and the other 10 containing 9406 form the testing data set. The artefact rates for the training and testing data are 31.6% and 30.5%, respectively. Each EEG record was normalised to be with zero mean and unit variance. In these experiments we used 5 fold cross-validation.

In an addition to the standard neural network technique, for comparison we used the evolutionary technique of GMDH, mentioned in Section 1, as well as the standard decision tree (DT) technique [23]. All the techniques listed in Table 3 ran 30 times for each fold in order to find the classifiers providing the best performance on the validation data. For each run the parameters of the classifiers were initialised randomly. The performances of the best classifiers on the validation data are evaluated on the testing data, and Table 3 provides their average values and standard variances calculated over the 5 folds.

**Table 3:** The average performance of the GMDH, DT, FNN and ECNN on the testing data

| # | Method | Perform, % | Variance |
|---|--------|-----------|----------|
| 1 | *GMDH* | 79.4 | 0.0044 |
| 2 | *DT* | 78.8 | 0.0128 |
| 3 | *FNN* | 79.4 | 0.0032 |
| 4 | *ECNN* | **79.8** | 0.0083 |

The GMDH, using the evolutionary search strategy, can learn polynomial networks of a near optimal complexity from data. The transfer function of neurons is assigned to be a short-term non-linear polynomial

$$y = w_0 + w_1 u_1 + w_2 u_2 + w_3 u_1 u_2, \tag{8}$$

where $w_i$ are the coefficients or synaptic weights and $u_1$ and $u_2$ are the inputs of neuron.

The above coefficients are fitted to the training data by using the standard least square method as described in [10, 12, 24]. It is important to note that, because the class boundaries in the EEG data overlap heavily, the fit of the coefficients to randomly selected training examples provides a better performance on the validation data. In our experiments we found that the best performance is achieved by selecting 50% of all the training examples.

The GMDH algorithm used in our experiments generates an initial population consisting of single-input neurons. Their transfer function is described as $y = w_0 + w_1 u_1$. Then the algorithm mates two randomly selected neurons and adds the offspring to the population if its performance on the validation data set becomes better, that is

$$p > \max(p_i, p_j),$$

where $p$ is the performance of offspring, $p_i$ and $p_j$ are the performances of parent-neurons randomly selected from the current population.

The above selection rule is a kind of local "elitist" selection described in [10, 12]. Here $i \neq j$, $i \in (1, N_r)$, where $N_r$ is the size of population of neurons of the $r$th generation. The number of offspring-neurons created for each generation is predefined.

The GMDH-type network is evolved while one or more created offspring-neurons improve the performance. When the number of failed attempts of improving the current performance is exceeded, the GMDH algorithm stops and selects a resultant network providing the best performance on the validation data. As the best performance can be achieved by several networks, the resultant network is assigned to be consisting of a minimal number of neurons.

In our experiments the initial population was assigned to be consisting of $N_1$ single-input neurons, here $N_1 = m = 72$. The number of offspring-candidates and the number of serial failed attempts were predefined to be 500 and 5, respectively.

The average performance of the above GMDH technique on the testing data and the standard variance were 79.4% and 0.0044 as shown in Table 3. The average number of two-input neurons with transfer function (8) and their variance were 25.2 and 3.63, respectively. The average number of the input variables involved in the GMDH-type networks was 20.8.

The DT technique used in our experiments exploits the randomised technique of assigning the partitions [22, 23]. This technique enables to prevent the DTs from over-fitting as follows. Searching the best partition at the $k$th level, the DT algorithm assigns the threshold $q_i^{(k)}$:

$$q_i^{(k)} \sim U(x_{i,\min}^{(k)}, x_{i,\max}^{(k)}), i = 1, ..., m,$$

where $x_{i,\min}^{(k)}$ and $x_{i,\max}^{(k)}$ are the minimal and maximal values of the data points of the $i$th variable which are available at the $k$th level partition, and $U$ is a uniform distribution.

The assignments of $q_i$ repeat a given number of times, say $n_s$, and the information gain is calculated for each $q_i$. The maximal value of information gain over $n_s$ is stored for each variable $x_1$, ..., $x_m$. Then the partition with the maximal information gate over all $m$ variables is assigned to be the best.

The partitions of the training data are repeatedly made while the number of data points exceeds a predefined number. This number can be preset as a fraction $p_{min}$ of the total amount of the training data.

In our experiments the DTs with the values of $n_s$ and $p_{min}$ equal to 25 and 0.06, respectively, has provided the best performance on the testing data. All these DTs consist of 4 nodes which involve correspondingly 4 input variables. Their average performance was 78.8% and the variance was 0.0128 as shown in Table 3. So, the GMDH-type neural network outperforms the DTs on average on 0.7%.

The average performance of the FNNs was 79.4%, and their variance was 0.0032. The best performance of the FNNs has been achieved for 19 principal components and 6 hidden neurons. These components contributed at least 99.5% to the classification outcomes. The performances of the GMDH-type neural network and the FNN, as we can see from Table 3, are the same.

When we applied the ECNN technique to these data, the average performance was 79.8% and the variance was 0.0083, as shown in Table 3. This is slightly better than that for the GMDH and FNN techniques. As we can see, the performance is worse than that in the first experiment in which we used only two EEG records labelled by one EEG-expert. This can happen because in our second experiments we used 30 EEGs recorded in different conditions and labelled by several experts. Moreover the artefact rate in these EEGs was higher.

It is also interesting to note that the average number of neurons involved in the ECNNs was equal to 7. This number is the same that has been found to be the best for the FNNs. However, the ECCNs involve on average 8 input variables selected from the original 72 variables. So, the ECNN technique seems to be better than the GMDH and FNN techniques.

## 7. Conclusion

We have developed a new algorithm allowing the cascade neural networks evolve and learn in the presence of noise and redundant features. The ECNN starts to learn with one neuron, and new inputs as well as new neurons are added to the network while its performance increases. As a result, the ECNN has a near optimal complexity.

The ECNN was applied for recognising artefacts in the clinical EEGs recorded from newborns during sleep hours. The artefacts in these EEG were visually labelled by EEG-viewers. Some of the spectral and statistical features, calculated to present the EEGs for an automated recognition, were noisy and redundant.

In our experiments with the artefact recognitions in the EEGs, the ECNNs, learnt from the data, has slightly outperformed the standard FNNs with a fixed structure. The ECNN has also outperformed the evolutionary GMDH-type as well as the standard decision tree techniques.

Thus, we conclude that the ECNN technique can be effectively used to train cascade neural networks in the presence of noise and redundant features. We believe that the ECCN technique can improve the performance of cascade neural networks applied to real-world problems of the pattern recognition and classification.

## Acknowledgments

This research has been supported by the University of Jena (Germany) and in part by the EPSRC, grant GR/R24357/01. The author is grateful to Frank Pasemann and Joachim Schult for fruitful and enlightening discussions, to Joachim Frenzel and Burghart Scheidt from the Pediatric Clinic of the University of Jena for providing the EEG recordings, and to Jonathan Fieldsend from the University of Exeter for useful comments.